\documentclass[final,5p,times,twocolumn]{elsarticle}

\usepackage[utf8]{inputenc}
\usepackage[T1]{fontenc}
\usepackage{algorithm}
\usepackage{algpseudocode}
\usepackage{amsmath,amssymb,amsfonts}
\usepackage{float}
\usepackage{graphicx}
\usepackage{stmaryrd} %another package with the common cause symbol
\usepackage{booktabs}
\usepackage{mathtools}

\DeclareMathOperator*{\argmin}{arg\,min}
\newcommand{\hcc}{\mathrel{\scalebox{0.8}{$\curlyveeuparrow$}}}

\journal{Neural Networks}

\begin{document}

\begin{frontmatter}



\title{Inference of hidden common driver dynamics by anisotropic self-organizing neural networks}


\author[inst1]{Zsigmond Benkő}

\affiliation[inst1]{organization={Department of Computational Sciences, HUN-REN Wigner Research Centre for Physics},
            addressline={Konkoly-Thege Miklós út 29-33}, 
            city={Budapest},
            postcode={1121},
            country={Hungary}}

\author[inst1]{Marcell Stippinger}
\author[inst1,inst2]{Zoltán Somogyvári}

\affiliation[inst2]{organization={Axoncord Ltd.},
            addressline={Dunakeszi u. 23}, 
            city={Budapest},
            postcode={1048},
            country={Hungary}}

\begin{abstract}
  We are introducing a novel approach to infer the underlying dynamics of hidden common drivers, based on analyzing time series data from two driven dynamical systems. The inference relies on time-delay embedding, estimation of the intrinsic dimension of the observed systems, and their mutual dimension. A key component of our approach is a new anisotropic training technique applied to Kohonen's self-organizing map, which effectively learns the attractor of the driven system and separates it into submanifolds corresponding to the self-dynamics and shared dynamics.

To demonstrate the effectiveness of our method, we conducted simulated experiments using different chaotic maps in a setup, where two chaotic maps were driven by a third map with nonlinear coupling. The inferred time series exhibited high correlation with the time series of the actual hidden common driver, in contrast to the observed systems. The quality of our reconstruction were compared and shown to be superior to several other methods that are intended to find the common features behind the observed time series, including linear methods like PCA and ICA as well as nonlinear methods like dynamical component analysis, canonical correlation analysis and even deep canonical correlation analysis.
\end{abstract}

\begin{graphicalabstract}
\includegraphics[width=1.5\linewidth]{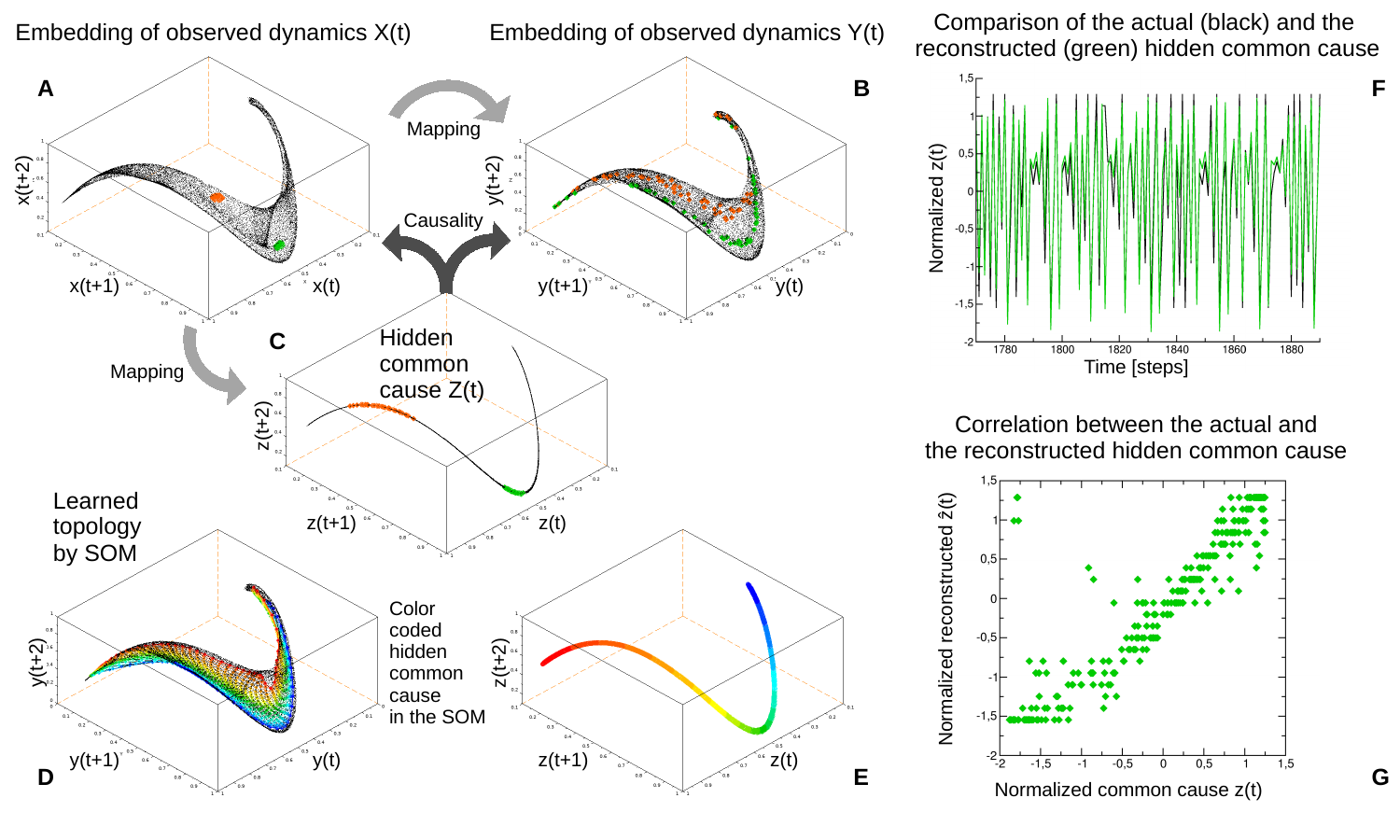}
\end{graphicalabstract}

\begin{highlights}
\item Hidden common driver time series are inferred from the driven dynamical systems

\item Integration of nonlinear dynamical systems theory with unsupervised learning 

\item A new anisotropic self-organizing map to learn manifold structures

\item More precise inference of hidden common variables than many established methods
\end{highlights}


\end{frontmatter}


\section{Introduction}\label{sec:intro}
The detection of causal relationships between observations holds significant importance in both scientific and everyday contexts. In the realm of science, various methods have been developed to uncover directed causal connections between observed time series, drawing from disciplines such as statistics, information theory, and the theory of nonlinear dynamical systems \cite{Granger1969, Pearl2000, Sugihara2012, Runge2018, Glymour2019, Assaad2022}. 

One of the most challenging problems in causal discovery is the potential existence of hidden confounders. Hidden confounders typically lead to false positive detections of direct causal connections in the majority of causal discovery methods. For this reason, several methods have been developed to mitigate the effects of hidden confounders \cite{Assaad2022}. These methods generally assume that the effects of a confounder are instantaneous between two observations or that they cause linear correlations between the observed data series \cite{ChuGlymour2008}. Additionally, these methods assume the absence of bidirectional connections or causal loops among the observed variables. In this context, a bidirectional arrow between variables is often interpreted as an indication of a hidden confounder.

While several nonlinear approaches to causality analysis have been explored, only a limited number have attempted to differentiate between direct causal connections and the influence of a hidden common driver without restricting the effect to be instantaneous and linear \cite{Hirata2010a, Benko2018complete, Benko2024}. The Dimensional Causality (DC) method \cite{Benko2018complete, Benko2024} allows for the identification of a hidden common driver by analyzing the estimated intrinsic dimensions of observed systems. Our novel approach, presented here, goes beyond mere detection by enabling the reconstruction of its time series.

Our main assumption is that the observed time series is predominantly generated by deterministic, ergodic dynamical systems. In this scenario, Takens' embedding theorem provides valuable insights into the properties of deterministic dynamics. It allows us to reconstruct the attractor of a dynamical system using only a single, scalar-valued time series observation, up to a smooth transformation that preserves the topology \cite{Takens1981}.

Furthermore, Jaroslav Stark extended Takens' theorem to unidirectionally connected dynamical systems and termed it the Forced Takens Theorem \cite{Stark1999}. He demonstrated that the forced system can be regarded as an observation of the forcing system. Therefore, embedding the forced system constitutes a reconstruction of both the forcing (driver) and the forced (driven) dynamics, capturing both the cause and the consequence. While Takens' theorem requires the observation function to be a smooth function of the system's variables, Stark's theorem additionally requires the connection between the two dynamical systems to possess the same smoothness.

This property has been effectively utilized by Sugihara's cross-convergent mapping (CCM) procedure, enabling the detection of unidirectional as well as circular connections between dynamical systems \cite{Sugihara2012}. While theoretically, CCM can determine the direction of causality without assuming an observable time delay between the cause and the consequence, the method has also been expanded to explicitly incorporate the time delay into the analysis \cite{Ye2015}, and extended to infer frequency-dependent causal information \cite{Benko2024a}. In the field of neuroscience, the initial application of CCM found that considering this time delay is necessary in certain cases to draw appropriate conclusions from the observations \cite{Benko2019}.

Stark's theorem \cite{Stark1999} establishes that the dynamics of a cause can be reconstructed by solely observing the consequence, as it forms a component of its reconstructed attractor. This fundamental property inspired the development of the DC analysis method \cite{Benko2018complete,Benko2024}, which aims to identify and quantify the probability of all possible types of causal relationships between two time series: independence, direct or circular causal connections, and the presence of a hidden common cause.

The DC method relies on the subadditivity of the system's attractor dimensions, with particular emphasis on the dimension of the joint attractor of the two systems. It has been demonstrated that the relationships between the joint and individual dimensions unequivocally determine the causal connections between the dynamical systems. By employing Bayesian inference, the DC method provides probabilities for different causal relationships. The effectiveness of the DC method was validated using simulated examples of "classic" chaotic dynamical systems, including nonlinearly coupled logistic functions, coupled Lorentz systems, and Hindmarsh-Rose models. Moreover, the DC method was applied to EEG data, where it was able to detect the effect of visual stimulation as well as the causal relationships between brain areas during epileptic seizures. This underscores the applicability of the DC method in localizing seizure onset zones \cite{Benko2018complete,Benko2024}. Notably, the DC method not only detects the existence of a hidden common driver but also provides an estimate of the dimension of its dynamics, which will be a key component in our work.

\subsection{Inferring Hidden Common Variables}

While the existence of a hidden confounder poses challenges for causal analysis, in many cases, these hidden variables that affect all observed variables are of central interest. For example, low-dimensional manifolds play a key role in population neural dynamics in the motor cortex \cite{Gosztolai2024}. Consequently, numerous methods have been developed to infer hidden common variables based on the observed ones.

The majority of these methods, such as Principal Component Analysis (PCA), Independent Component Analysis (ICA), Dynamical Component Analysis (DCA), and Canonical Correlation Analysis (CCA), assume that the observed variables are linear combinations of the hidden ones. These methods then identify these hidden linear combinations by maximizing different measures.

To overcome the limitations of linear models, some of these methods have been extended to nonlinear approaches. For instance, Deep Canonical Correlation Analysis \cite{Andrew2013} uses a feed-forward deep neural network to map the observed subsystems to 1D subspaces, where the correlation between the values is maximized.

Slow Feature Analysis (SFA) is also intended to reveal common features between observed time series. SFA applies a nonlinear expansion of the feature space followed by a linear projection to find the common feature with the smoothest temporal dynamics \cite{Wiskott98, Wiskott2002}.

On the other hand, the assumption that the observed signals are generated by interconnected dynamical systems has led to methods where linear mixing is not required.

The initial attempt to reconstruct the dynamics of a hidden common driver was undertaken by Timothy Sauer \cite{Sauer2004}. Recognizing that the dynamics of a hidden common driver constitute a shared component within the observed dynamics of the driven systems, Sauer proposed the possibility of determining this shared component through the observation of multiple systems, utilizing Takens' embedding theorem. Sauer's method \cite{Sauer2004, Sauer2010} involved decomposing one of the observed and reconstructed driven attractors into equivalence classes corresponding to fixed values in another driven attractor. This approach was demonstrated in systems where the observed attractors were at most one-dimensional manifolds, such as chaotic logistic maps driven by periodic dynamics on discrete points or periodic oscillators driven by other discrete periodic oscillators.

More recently, William Gilpin \cite{Gilpin2023} advanced Sauer's method. His approach involves calculating the Laplacian of the shared adjacency matrices among the driven systems and utilizing topological ordering within the equivalence classes.

In contrast, our approach takes a different path. By conducting a dimensional analysis of the driven systems, we establish the foundational topology of the driver-driven joint system. Subsequently, we employ a neural network model that conforms to the predefined topological structure of the observed manifolds and train it on the observed time series in an unsupervised manner.

\subsection{Self-organizing maps}

A crucial tool for achieving our objective is a modified version of Kohonen's self-organizing map (SOM) \cite{Kohonen1982}. A self-organizing map is a neural network model that exhibits a distinctive form of unsupervised learning. Through this learning process, the network can adapt its pre-defined neural structure, characterized by the connections between neurons, to the input manifold, which is determined by the input samples in input space. Consequently, the network provides a topological representation, or map, of the input space.

The fundamental structure of a SOM consists of a grid of neurons. While typically a 2D grid is used, 1D or higher dimensional grids can also be employed. Each neuron in the grid is characterized by a normalized synaptic weight vector that connects the input space to the neuron, essentially marking the center of its receptive field in the input space.

During the learning process, input samples are presented sequentially. The activation of a neuron at each presentation depends on the distance between the actual sample and the neuron's receptive field center, i.e., its weight vector. A winner-takes-all mechanism is introduced, fostering competition among activated neurons. The neuron whose receptive field center is closest to the actual sample wins the competition and drives the actual learning step. All neurons' synaptic weights adjust slightly towards the actual sample, but the learning rate varies based on the distance of each neuron from the winner along the inner neural grid. As a result, the synaptic weight of the winner is modified by the most significant amount, while the weight modification decreases the farther a neuron is from the winner.

This competition among neurons for input samples leads to an even distribution of receptive field centers along a homogeneous input manifold. When dealing with inhomogeneous samples, the density of receptive field centers is proportional to the density of the input space. Moreover, neurons that are located closely in the inner grid also tend to have receptive fields that occupy nearby positions in the input space. As a result, the grid forms a smooth mapping of the input manifold.

In our framework, the SOM is trained using not just one, but two sets of inputs derived from the embedding of the two driven systems. The original training method has been modified to enable anisotropic map fitting. This anisotropic fitting facilitates the decomposition of the observed attractor into submanifolds that describe the self-dynamics of the system as well as dimensions that correspond to the dynamics of the hidden common driver. By incorporating this modified SOM, we can effectively unravel the underlying dynamics and separate the contributions of the self-dynamics and the hidden common driver.

\section{Methods}

\subsection{Simulations}

Two types of discrete-time, continuous-space, one-dimensional chaotic dynamical systems were simulated to generate time series for analysis: logistic maps and tent maps. The basic form of the logistic map is:

\begin{equation}
x(t+1) = r x(t)\left(1 - x(t)\right)
\label{eq:logist}
\end{equation}

To demonstrate the topological relationships between driver and driven dynamical systems, we simulated unidirectionally coupled logistic maps:

\begin{equation}
\begin{split}
x(t+1) &= r_1 x(t)\left(1 - x(t)\right) \\
y(t+1) &= r_1 y(t)\left(1 - y(t) - \beta_1 x(t) \right).
\end{split}
\label{eq:unidirlogist}
\end{equation}

Here, we used the parameter $r_1=3.86$ and a connection strength $\beta_1=0.5$.

For the case of circular nonlinear coupling, our example dynamical system is defined as follows:

\begin{equation}
\begin{split}
x(t+1) &= r_2 x(t)\left(1 - x(t) - \beta_2 y(t)\right) \\
y(t+1) &= r_2 y(t)\left(1 - y(t) - \beta_3 x(t) \right).
\end{split}
\label{eq:circlogist}
\end{equation}

In this system, we used the parameter $r_2=3.86$ and connection strengths $\beta_2=0.5$ and $\beta_3=0.6$.

The training process was demonstrated on a triad of coupled logistic systems:

\begin{equation}
\begin{split}
z(t+1) &= r_3 z(t)\left(1 - z(t)\right) + \eta_1(t) \\
x(t+1) &= r_3 x(t)\left(1 - x(t) - \beta_4 z(t)\right) + \eta_2(t) \\
y(t+1) &= r_3 y(t)\left(1 - y(t) - \beta_5 z(t) \right) + \eta_3(t)
\end{split}
\label{eq:hcclogist}
\end{equation}

Here, we used the parameter $r_3=3.8$ with connection strengths $\beta_4=0.4$ and $\beta_5=0.3$. Additionally, a small amount of Gaussian noise was added to all the variables: $SD(\eta_1)=SD(\eta_2)=SD(\eta_3)=0.001$.

For systematic evaluation and comparison with different methods, simulations were repeated 50 times by randomly selecting the $r$ parameters of the three logistic maps independently from a uniform distribution in the range [3.8, 4], the $\beta$ coupling parameters from the interval $[0.1, 0.5]$, and the initial conditions for $x(t)$, $y(t)$, and $z(t)$ from the interval [0, 1]. Time series of 20,000 steps were generated without noise, with 10,000 steps used for training the SOM and 10,000 steps for testing.

The basic update rule for the tent map is given by the equation:

\begin{equation}
x(t+1) = T(x(t)) = 1 - 2|x(t)-1/2|
\label{eq:tent}
\end{equation}

where $x \in (0, 1) \subset \mathbb{R}$. However, for our simulations, we used a modified, tilted map:

\begin{equation}
\begin{split}
x(t+1) &= T_{\alpha}(x(t)) = \frac{1}{2\alpha(\alpha-1)}|x(t)-\alpha|
\\
&+\left(\frac{1}{\alpha}+\frac{1}{2\alpha(\alpha-1)}\right)\left(x(t)-\alpha\right)+1
\end{split}
\label{eq:tentalpha}
\end{equation}

The peak of this triangular map is 1, but it is now positioned at $\alpha$.

A coupled triad of tent maps was simulated for systematic performance evaluation:

\begin{equation}
\begin{split}
z(t+1) &= T_{\alpha_1}(z(t))\\
x(t+1) &= T_{\alpha_2}(x(t)+\beta_6 z(t))\\
y(t+1) &= T_{\alpha_3}(y(t)+\beta_7 z(t))
\end{split}
\label{eq:hcctent}
\end{equation}

The $\alpha$ parameters were selected from the range $[0.1, 0.5]$, the $\beta$ connection strengths were chosen from the interval $[0.1, 1]$, and the initial conditions were set within the interval [0, 1]. Time series of 20,000 steps were generated without noise, with 16,000 steps used to train the SOM and 2,000 steps to evaluate the correlation.

\subsection{Embedding and cross-mapping}

The topological relations between the observed systems are demonstrated on the reconstructed attractor manifolds by time delay embedding, according to Takens' embedding theorem:
\begin{equation}
\begin{split}
	X(t) &= [x(t), x(t + \tau), \dots, x(t + (m-1) \tau)]\\
    Y(t) &= [y(t), y(t + \tau), \dots, y(t + (m-1) \tau)]
\end{split}
\label{eq:Takens}
\end{equation}
In the specific examples, we set $\tau=1$ time-step and $m=3$, resulting in the embedding of all time series into three dimensions. To distinguish between scalar-valued and vector-valued time series, lowercase letters are used for the former, while capital letters are employed for the embedded vector-valued time series.

 Convergent Cross-Mapping (CCM) \cite{Sugihara2012} is not able to distinguish the effect of a hidden common driver from the direct causality, but was utilized here to demonstrate the topological relations between manifolds. Thus, cross-mapping from system $X$ to system $Y$ is conducted similarly to \cite{Sugihara2012}: 
\begin{itemize}
    \item Initially, a random state $X(t_s)$ is selected from system $X$.
    \item Next, we search for the $K$ nearest neighbor states $X(t_K)$ in the reconstructed state space of system $X$, in terms of Euclidean distances.
    \item Finally, the corresponding states $Y(t_K)$ from system $Y$ are retrieved.
\end{itemize}

\subsection{Dimensional Causality Analysis}

The first step in our analysis pipeline is to perform Dimensional Causality analysis \cite{Benko2018complete, Benko2024} on the two observed time series. This analysis assigns probabilities to the five basic possible causal relationships between the two observed time series, including the presence of a hidden common driver between otherwise independent systems. If DC analysis assigns a high probability to the existence of a hidden common driver, the analysis method based on the SOM, as presented here, can be used to infer that hidden common driver.

\subsection{Dimensions of the SOM}

The second step of the analysis is to construct the SOM with proper dimensions. To properly construct SOM, it is essential to have knowledge about the dimensions of both the observed and hidden common driver dynamics. Within the Dimensional Causality framework, estimates of the embedded manifolds are calculated by a K-nearest neighbor-based method proposed by \cite{Farahmand2007} and modified by \cite{Benko2020a}.

In order to estimate the intrinsic dimension of the embedded time series $X(t)$, we iterate through all the points and search for its $k^{th}$ and $2k^{th}$ nearest neighbors. Let the distance of the $k^{th}$ neighbor from the original $X(t)$ point be denoted as $R_k(t)$. Then, the local dimension $d_k(X(t))$ around $X(t)$ is estimated as follows:
\begin{equation}\label{eq:local_fsa}
    d_k(X(t)) = \frac{\ln(2)}{\ln \left( R_{2k}(t)/R_{k}(t) \right)}
\end{equation}

While Farahmand et al.\,\cite{Farahmand2007} used the mean of the local dimension estimators as a global estimate, Benkő et al.\,\cite{Benko2020a} showed that the median of the local dimensions provides an unbiased global estimate of the manifold dimension:
\begin{equation}
D_{X}=Median(d_k(X(t)))
\end{equation}

The intrinsic dimension $D_Y$ of time series $Y$ is estimated in a similar manner.

To determine the optimal neighborhood parameter $k$, two factors must be considered: higher $k$ values enhance the robustness of dimension estimation against noise, but they also introduce more bias if the manifolds are highly curved. In our case, dimension estimations were averaged over the interval $k \in [10,20]$, where the estimates remained relatively stable. This range was effective given the actual noise level and the high curvature of the manifolds.

The dimension of the hidden common driver, denoted as $D_Z$, can be estimated by calculating the Mutual Information Dimension between the time series $x(t)$ and $y(t)$ \cite{Sugiyama2013}. This involves comparing the dimensions of the two observed systems and the dimension of their joint system, denoted as $D_J$:
\begin{equation}
D_{Z}=D_X+D_Y-D_J
\label{eq:DZ}
\end{equation}

While in \cite{Sugiyama2013} the joint manifold $J$ is a direct product of the $X$ and $Y$ manifolds, thereby residing in an embedding space of double dimension:
\begin{equation}
J(t)=[X(t),Y(t)],
\end{equation}
in the DC framework, the joint attractor is created by embeddig a joint observation function:
\begin{equation}
J(t)=X(t)+aY(t),
\end{equation}
where $a=\sqrt{29/31}$ is a properly chosen irrational number. In this framework, the joint attractor resides in a space with the same dimensions as the individual observations.

Dynamical systems, especially chaotic ones, often result in fractal attractors with non-integer dimensions. Since the dimension of the SOM is an integer, the closest integer should be used to define the dimension of the SOM.

In our demonstration example, both the hidden common cause and the driven consequences are described by logistic dynamics. As a result, the hidden common driver exhibits dynamics with a dimension close to one, while both driven systems exhibit dynamics with dimensions close to two. Therefore, a 2D anisotropic grid is necessary to fit the observed dynamics, with one of these two dimensions representing the dynamics of the hidden common driver.

The anisotropic SOM is built up from a 2D grid of $40\times20$ nodes. Each node is characterized by its position in the 2D grid described by indices $i$ and $j$ and the center of its receptive field in the reconstructed $Y$ space, described by a 3D vector $C_{ij}$. The aim of the training procedure is to fit the $C_{ij}$ points evenly to the manifold $Y$ so that the 2D grid forms a discretized topographical map of the 2D manifold $Y$. Furthermore, we require an anisotropic fitting, so that the first (40-node long) dimension follows the submanifolds corresponding to the self-dynamics of the system $Y$, while, the second (20-node long) dimension corresponds to the different states of the hidden common cause. Each self-dynamics submanifold is formed by those points, among which the system $Y$ moves, while the value of the hidden common cause $Z$ is fixed. Thus, the movement of the state of the system along these submanifolds is determined by the self-dynamics, while moving between submanifolds is driven by the dynamics of the hidden common cause.

In general, the SOM should have $D_Y$ dimensions to effectively learn the $Y$ manifold. Among these dimensions, $D_Z$ describe the common driver dynamics, while the remaining $D_Y - D_Z = D_J - D_X$ dimensions represent the self-dynamics of the system $Y$.

\subsection{Anisotriopic training of the SOM}

$T=10000$ time steps long simulated time series were used to train the SOM network.
Initially, the $C_{ij}$ vectors were distributed randomly with uniform distribution within the unit cube.  
The anisotropic training consisted of two nested loops: The outer loop run $N=10000$ times. In this loop, a random seed element $X(t_s)$ of the $X$ manifold is chosen and mapped to the corresponding $Y(t_s)$ point. The activation of the nodes depends on the distance between the presented $Y(t_s)$ sample and their receptive field centers $C_{ij}$. Then the most activated overall winner node $C_{i^*j^*}$ of the SOM was chosen that center is the closest to $Y(t_s)$:
\begin{equation}
i^*,j^*= \argmin_{ij}(||C_{ij}-Y(t_s)||)   
\end{equation}

In the next step, the closest $X(t_K)$ neighbors ($K=20$) are searched for and mapped to the corresponding $Y(t_K)$ elements. The $Y(t_K)$ points are formed as a one-dimensional submanifold in the $Y$ space.

The second loop starts here and the elements of the $Y(t_K)$ are presented to the SOM one after another.
The training procedure becomes anisotropic at this point: The distance between the actual $Y(t_k), k\in K$ and \emph{only the $C_{.,j^*}$ row of}  $C_{i^*j^*}$ is calculated, and the actual winner $C_{i^aj^*}$ is chosen according to the smallest distance:
\begin{equation}
i^a=\argmin_i (||C_{i,j^*}-Y(t_k)||)
\label{eq:winner}
\end{equation}
The learning weights are calculated for all nodes according to their anisotropic distances from $C_{i^aj^*}$:
\begin{equation}
w_{ij}(s)=e^{-((i-i^a)^2/\sigma_1^2(s)+(j-j^*)^2/\sigma^2_2(s))}
\end{equation}

Both $\sigma_1(s)$ and $\sigma_2(s)$ depend on the simulation steps done, resulting an anisotropic shrinkage of the learning area within the grid, during learning:
\begin{equation}
\begin{split}
	\sigma_1(s) &= 10\times e^{-s/N}\\
    \sigma_2(s) &= 20\times e^{-(s*log 5)/N}
\end{split}
\label{eq:sigma}
\end{equation}
As a last step within the second loop, the center of perception vectors of all grid nodes are updated:
\begin{equation}
C_{ij}(s+1)=C_{ij}(s)+\epsilon(s)\times w_{ij}\times (Y(t_k)-C_{ij}(s))
\end{equation}
Where learning rate $\epsilon(s)$ decreased with the simulation steps of the outer loop, resulting in a stabilization of the centers by the end of the learning:
\begin{equation}
\epsilon(s)= 0.2\times e^{-(s*log 20)/N}
\end{equation}
After all the $Y(t_k), k\in K$ points are presented to the SOM, the second loop terminates, and a new round of the first loop startes by choosing a new $X(t_s)$ randomly. For a pseudocode description of the algorithm see Algorithm \ref{Algo1}. 

\begin{algorithm*}
    \caption{Anisotropic SOM training algorithm}
    \begin{algorithmic}
    \State $X \gets \Phi(x), Y \gets \Phi(y)$  \Comment{Time delay embedding of time series}
    \State $\gamma \gets 0$
    \State $C_{i, j}(\gamma) \gets r_{ij}$ \Comment{Initialize the grid}
    \For{$s = 1$ to $N$}
        \State Compute $\sigma_1(s), \sigma_2(s), \epsilon(s)$ \Comment{Setting learning rates}
        \State $t_s \gets r_s$ \Comment{Pick a random time index}
        \State $i^{\star}, j^{\star} \gets \argmin_{i,j} (||C_{i, j}(\gamma) - Y(t_s)||$) \Comment{Find nearest element on the grid}
        \State $t_{1}, \dots, t_{K} \gets$ KNN$(X(t_s))$  \Comment{Find the time index of nearest neighbors}
        \For{$k = 1$ to $K$}
            \State $i^a \gets \argmin_i |Y(t_{l^k}) - C_{i,j^{\star}}(\gamma)|$ \Comment{Find the nearest grid-point within the selected column}
            \State $w_{ij}(\gamma) \gets \exp{[-((i-i^a)^2/\sigma_1^2(s)+(j-j^*)^2/\sigma^2_2(s))]}$ \Comment{Compute weights}
            \State $C_{ij}(\gamma+1) \gets C_{ij}(\gamma)+\epsilon(s) w_{ij}(\gamma) (Y(t_k) - C_{ij}(\gamma))$\Comment{Update grid-point coordinates}
            \State $\gamma \gets \gamma+1$
        \EndFor
    \EndFor
    \end{algorithmic}
\label{Algo1}
\end{algorithm*}

\subsection{Readout}
To readout the estimated actual value of the hidden common driver from the well-trained SOM, we look at the second index of the winner node. For each time instance $t$ in the test set, the embedded input patterns $X(t)$ and $Y(t)$ are presented to the SOM, and the winner node is selected similarly to the training phase, according to Eq.\,\ref{eq:winner}. The second index $j^\star$ of the winner node provides a discretized estimate $\hat{Z}$ that represents a topologically smooth mapping of the hidden common driver. While we cannot directly obtain the real numerical values of the hidden common driver, we can transform the time series of the winner $j^\star$ indices into comparable values by applying a smooth transformation. Therefore, both time series are mean-corrected to zero and normalized by their standard deviations, assuming that scaling and shifting as smooth transformations are sufficient for comparisons. To assess the scale-independent comparison, we calculate the correlation between $Z(t)$ and $\hat{Z}$.

In summary, our workflow involves the following steps:

    Time delay embedding of the two observed signals ($x(t)$ and $y(t)$) resulting in two manifolds, created by the vector-valued time series $X(t)$ and $Y(t)$.
    Creating the joint attractor by as the embedding of the joint observation $J(t)=X(t)+aY(t)$.
    Estimating the $D_X$, $D_Y$, and $D_J$ dimensions of the $X$, $Y$, and $J$ manifolds.
    Calculating the mutual dimension: $D_Z = D_X + D_Y - D_J$.
    Creating the anisotropic neural grid of the SOM, with $D_J - D_Y$ dimensions for the self-dynamics of system $X$ and $D_Z$ dimensions for the shared dynamics.
    Anisotropic training of the SOM on $X(t)$ and $Y(t)$ time series.
    Presenting the test set $X(t)$ and $Y(t)$ and reading out $\hat{Z}(t)$ from the trained SOM.

\section{Results}

\subsection{Cross-mapping}

\begin{figure*}[htb!]
\centering
\includegraphics[width=0.8\linewidth]{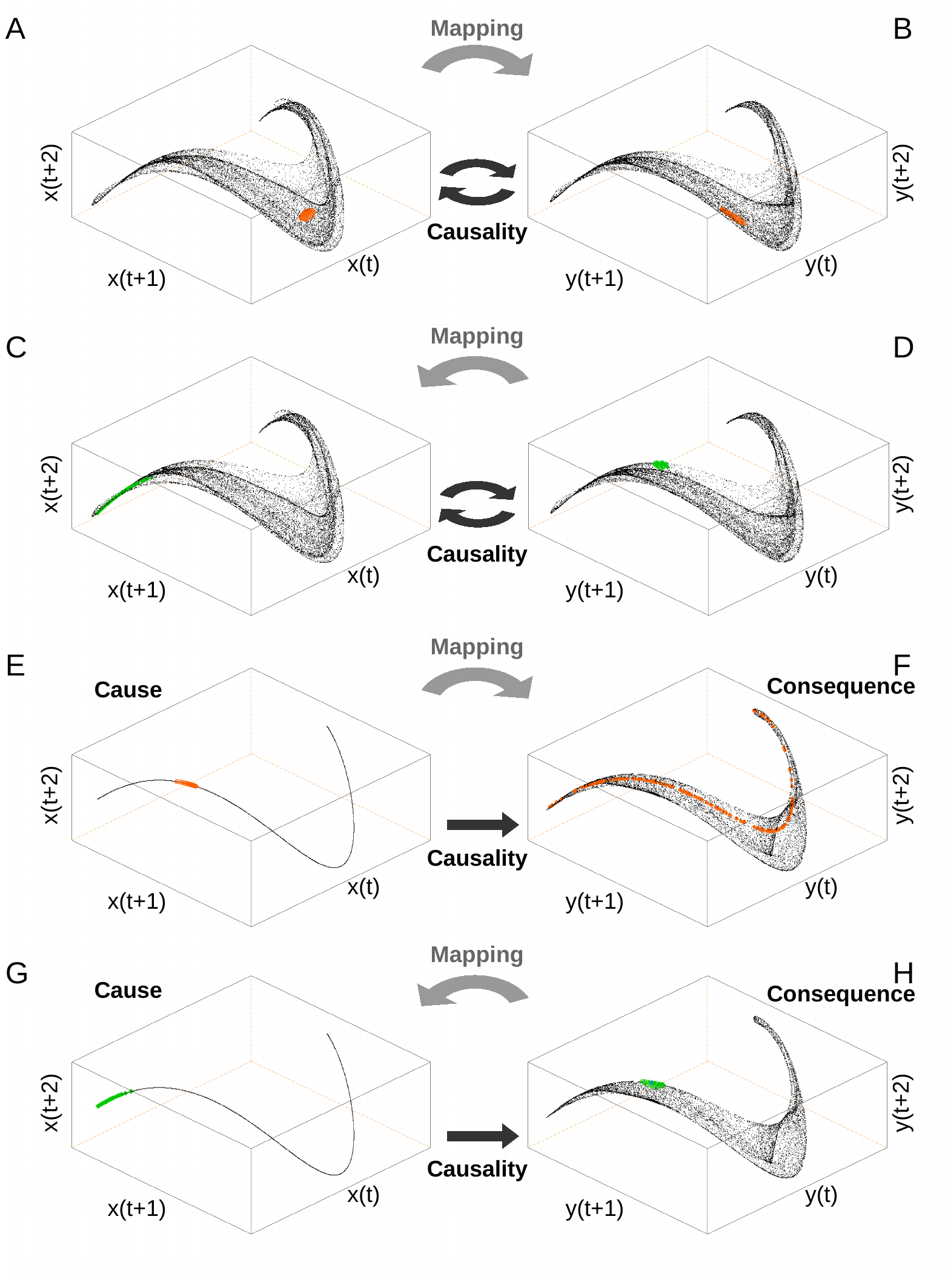}
\caption{\textit{Crossmapping between coupled logistic systems, represented in 3D embedding.} A-D: In the case of circular coupling between the systems, the mapping behaves similarly in both directions. A local neighborhood in the state space of system X (A) is mapped into a local neighborhood in the state space of system Y (B). Similarly, a local neighborhood in the state space of system Y (D) is mapped into a local neighborhood in system X (C). However, in unidirectional coupling (E-H), the dimensions of the two systems differ, resulting in different behavior for the projections in the two directions. A small local neighborhood in the cause (E) is mapped onto a one-dimensional submanifold in the state space of the consequence (F), while a small local neighborhood in the consequence (H) is mapped into a small local neighborhood in the cause (G).
}
\label{fig:CCM}
\end{figure*}

First, to gain insight into the topological relations between coupled dynamical systems, we demonstrate the properties of cross-mapping in two cases: First between circularly, second between unidirectionally coupled logistic maps (see Fig.\,\ref{fig:CCM}).

In the circularly coupled case (Eq.\,\ref{eq:circlogist}), the two reconstructed manifolds are topologically equivalent, their dimensions are equal (two in the actual case), and the cross-mapping works well in both directions (Fig.\,\ref{fig:CCM} A-D). The mapping of a small neighborhood around any points of the manifolds results in comparably small patches of neighboring points in the other manifold.

Next, we examined the properties of cross-mapping on unidirectionally coupled systems (Eq.\,\ref{eq:unidirlogist}). Here the dimensions of the two manifolds are different: the cause ($X(t)$) forms a 1D manifold in the embedding space, while the consequence ($Y(t)$) results in a 2D manifold in the 3D embedding space (Fig.\,\ref{fig:CCM} E-H). Due to this dimensional difference, the two manifolds can not be topologically equivalent: cross-mapping implements an injective but not a surjective function from $X$ to $Y$. While cross-mapping of a small neighborhood from $Y$ (Fig.\,\ref{fig:CCM} H) to $X$ results in a similarly small patch in $X$ (Fig.\,\ref{fig:CCM} G), cross-mapping a small neighborhood in the X manifold (Fig.\,\ref{fig:CCM} E) forms a 1D submanifold in the 2D manifold of $Y$ (Fig.\,\ref{fig:CCM} F). As the system $Y$ moves along these submanifolds at a given value of the cause $X$, these submanifolds describe a constituent of the $Y$ dynamics, that we called self-dynamics. Meanwhile, the cause drives the system $Y$ between these submanifolds, thus each submanifold represents a specific value of the cause $X$.

\begin{figure*}[htb!]
\centering
\includegraphics[width=0.8\textwidth]{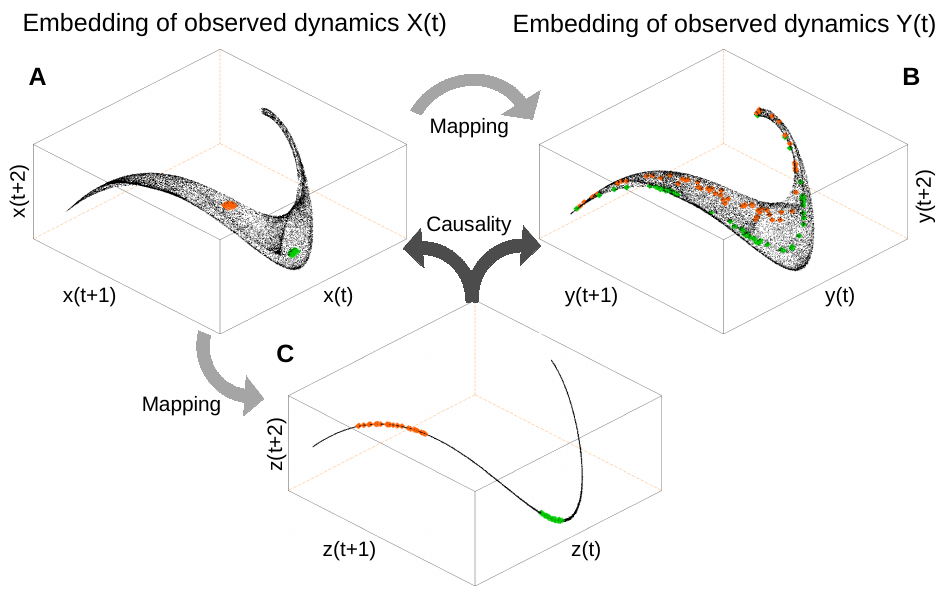}
\caption{\textit{Cross-mapping with a hidden common driver.} The Z system is the hidden common driver (C) that drives the two uncoupled systems X and Y on A and B respectively. Cross-mapping of the orange and green local patches forming two local neighborhoods in the state space of $X$ (A), results in two well-localized patches in the state space of the unobserved Z system (C), however mapping these local neighborhoods onto the $Y$ system results in two one-dimensional submanifolds (B).} 
\label{fig:HCC}
\end{figure*}

Now consider the case of the hidden common cause (Eq.\,\ref{eq:hcclogist}), where two observed systems (X, Fig.\,\ref{fig:HCC} A) and (Y, Fig.\,\ref{fig:HCC}, B) are both driven by the variable $Z$ of a third unobserved logistic map (Z, Fig.\,\ref{fig:HCC}, C). Both observed systems form two dimensional manifolds, similar to the circularly coupled case, but the crossmapping behaves differently.

Cross-mapping of any small patches from manifold $X$ (orange and green patches on Fig.\,\ref{fig:HCC}, A) results in small patches in the hidden cause Z (Fig.\,\ref{fig:HCC}, C) while the same patches are mapped to two different 1D submanifolds in the manifold $Y$ (Fig.\,\ref{fig:HCC}, B). Thus, similarly to the unidirectional case, the 2D manifold $Y$ can be decomposed into such 1D submanifolds, but now each submanifold corresponds to a given value of the hidden common cause $Z$. The self-dynamics moves the system $Y$ \emph{along} these submanifolds, while, the dynamics of the hidden common cause drives the system $Y$ \emph{between} these submanifolds.

\begin{figure*}[htb!]
\centering
\includegraphics[width=\textwidth]{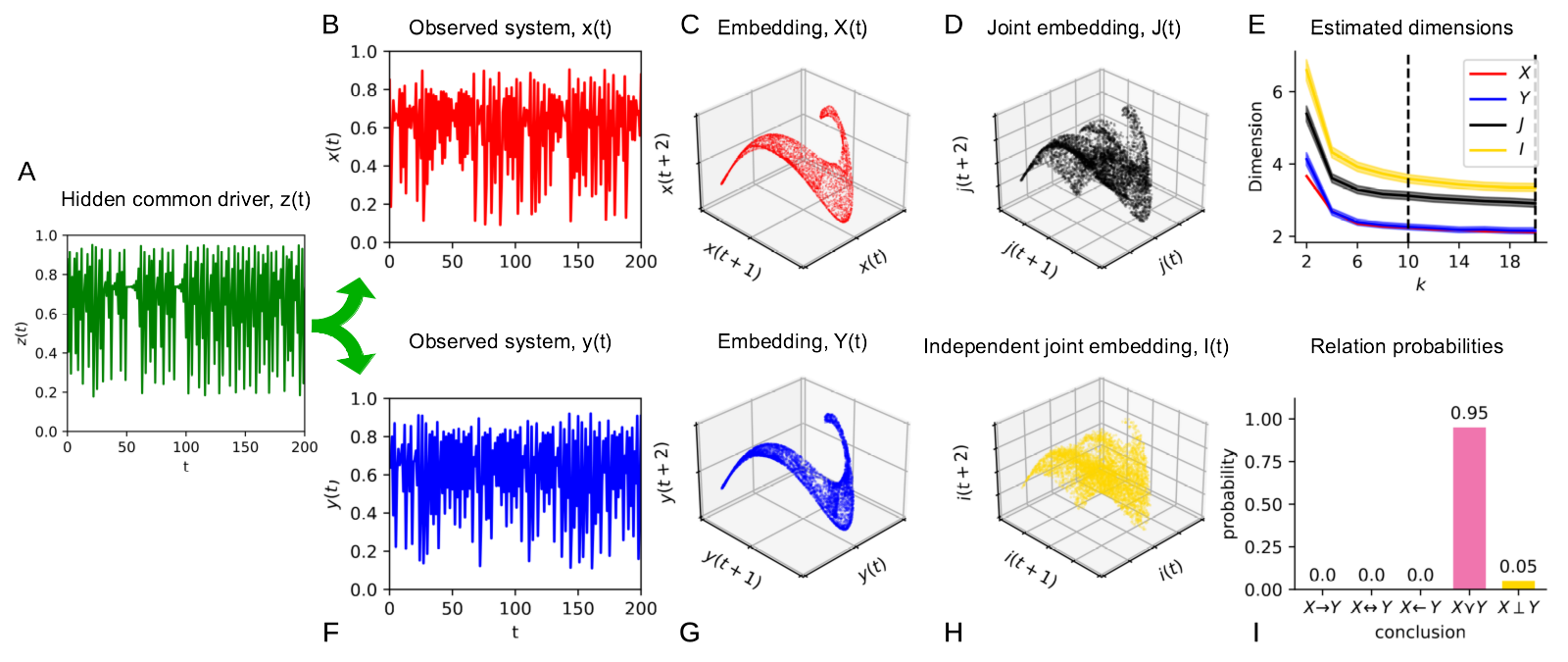}
\caption{\textit{Dimensional Causality analysis.} Three connected logistic map was simulated and analysed. A: time series of the hidden common driver, z(t), (green). B and F: time series of both observed time series (x(t), red and y(t) blue). C and G: the observed time series are embedded in 3 dimensions. D: Joint embedding (J(t), black). H: Time permuted joint embedding (I(t), yellow). E: Estimated dimensions as a function of the neighborhood size (k). The interval between 10 and 20 was used for the Bayesian estimation. I: Estimated posterior probabilities of the five basic causal relations: $X \rightarrow Y$: $X$ drives $Y$; $X \leftrightarrow Y$: bidirectional coupling; $X \leftarrow Y$: $Y$ drives $X$; $X \hcc Y$: a hidden common cause drives both; $X \perp Y$: $X$ and $Y$ are independent. The dimensional analysis assigns the highest probability to the existence of the hidden common driver.}
\label{fig:DC}
\end{figure*}

\subsection{Dimensional Causality analysis}

Before the inference of the hidden common driver actually started, the zeroth step was the Dimensional Causality (DC) analysis to investigate the existence of the hidden common driver and to measure the dimensions of the systems \cite{Benko2018complete, Benko2024} (Fig.\,\ref{fig:DC}). The steps of our analysis pipeline will be demonstrated on a simulated time series of a triad of coupled logistic maps, similar to that was presented in the previous example: the common driver (Fig.\,\ref{fig:DC} A, green, z(t)) drives the two observed systems (Fig.\,\ref{fig:DC} B, red x(t) and C, blue, y(t)). Within the DC analysis, a chunk of length 5000 steps of the observed time series were applied and embedded into 4 dimensions manifolds (Fig.\,\ref{fig:DC} D and E). The dimensions of both manifolds were estimated to be close to two: $D_X=2.17 \pm 0.05$ and $D_Y=2.19 \pm 0.04$ (mean and STD), while the joint dimension of the systems was $D_j = 3 \pm 0.07$ and the dimension of the independently time-permuted joint $D_I = 3.45 \pm 0.13$. The principle of the DC analysis states, that the relations between the $D_X$,$D_Y$, $D_J$ and $D_I$ dimensions unequivocally determine the causal relationship between the observed systems. Specifically, in the actual case:
\begin{equation}
Max(D_X,D_y) < D_J < D_I
\end{equation}
That implies that the X and the Y systems are unconnected, but there exists a hidden common driver behind them. Thus, the Bayesian evaluation procedure for the dimensional relations assigns high probability to the existence of the hidden common driver (Fig.\,\ref{fig:DC} F). For further details of the DC analysis the reader is referred to \cite{Benko2024}.

Based on the above results, the dimension of the hidden common driver could be estimated by the Eq.\,\ref{eq:DZ}. Thus $D_Z=D_X+D_Y-D_J = 1.36$, while the self dynamics of $X$ is $D_X-D_Z = 0.81$ and the self dynamics of $Y$ is $D_Y-D_Z = 0.83$ dimensional. These results are in accordance to the properties of the simulated systems and serves as a basis for setting the dimensions of the anisotropic SOM. As the SOM requires integer dimensions, both the self dynamics and the dynamics of the hidden common driver are rounded to 1, thus a 1+1 dimensional ASOM is set up to properly learn the attractor manifold structure of any of the observed systems.

\begin{figure*}[htb!]
\centering
\includegraphics[width=\textwidth]{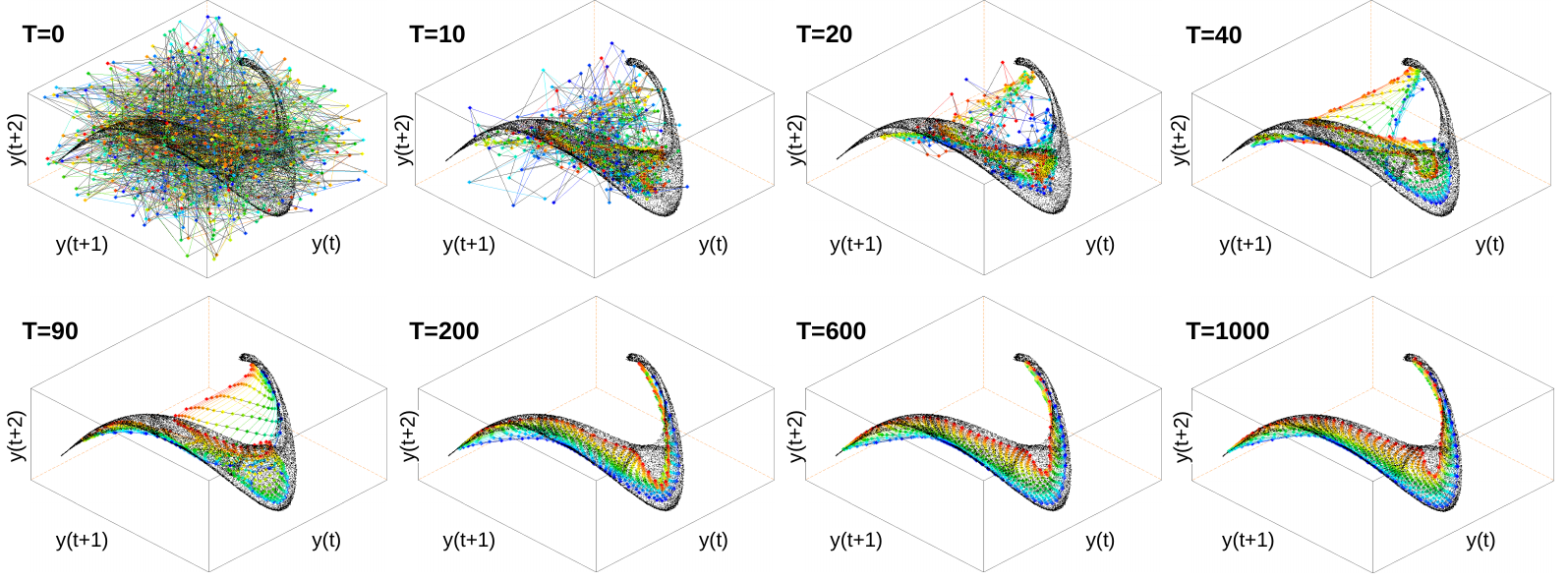}
\caption{\textit{Development of the SOM during anisotropic learning.} During each time step, a random point in manifold $X$ was chosen, and its local neighbourhood of $K=20$ points were presented to the network, forming a bundle in the manifold $Y$ marked by different colors. The $20 \cdot 40$ nodes of the grid represents the $C_{i,j}$ centers of the receptive fields of the neurons, while the edges corresponds to the predefined neighbourhood structure of the network.}
\label{fig:ASOMLearning}
\end{figure*}

\begin{figure*}[htb!]
\centering
\includegraphics[width=\textwidth]{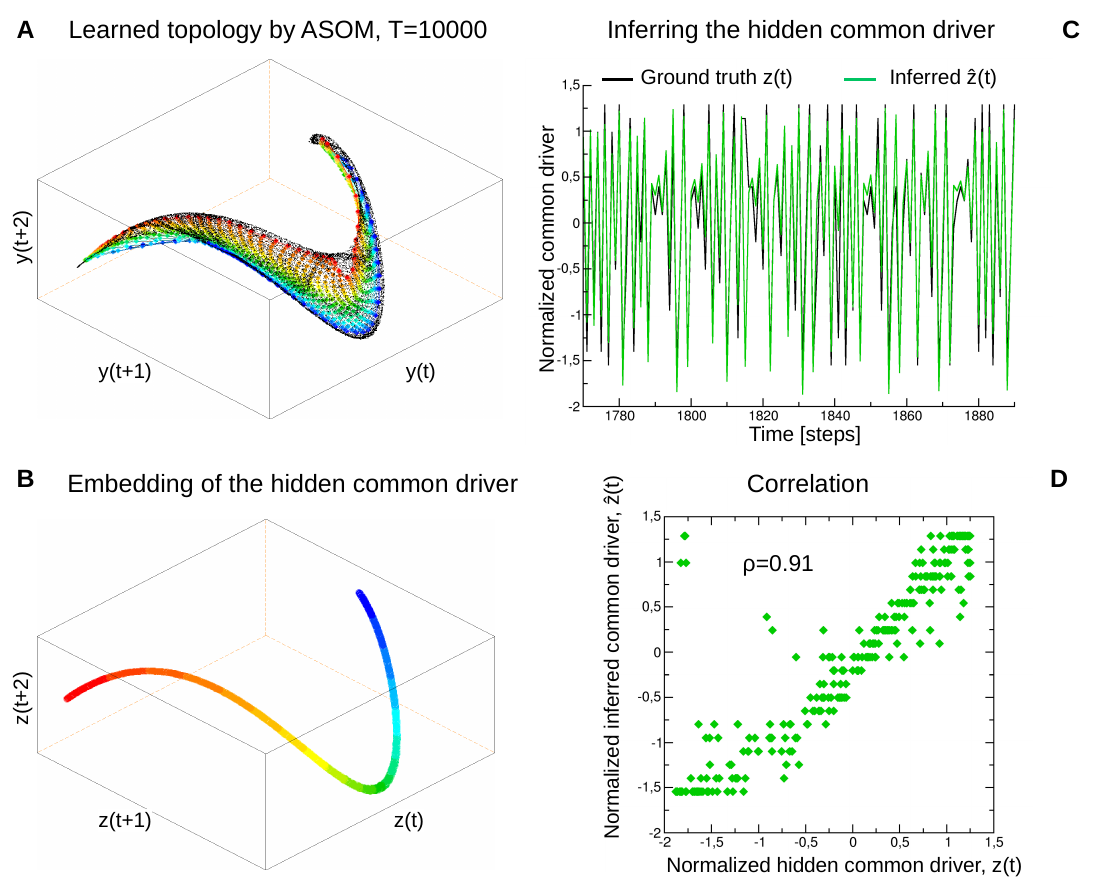}
\caption{\textit{Readout of the hidden common driver.} A: The trained 2D self-organizing map, after $N=10000$ training steps, fits well to the Y manifold and divides it into stripes marked by different colors. The nodes of the grid represent the $C_{i,j}$ centers of the receptive fields of the neurons. The self-dynamics of system Y move the system within the colored stripes on the SOM, while stripes of different colors correspond to the various states of the hidden common driver, which are marked by the same colors in B. C: Comparison of the actual and the reconstructed hidden common driver. Both time series are normalized by standard deviation (SD). D: Correlation between the actual and the reconstructed normalized hidden common driver values ($\rho=0.91$).} 
\label{fig:ASOMReadout}
\end{figure*}

\subsection{Training of the ASOM network}

The task of the anistropic training of the SOM network is to learn the decomposition of the attractor manifold shown in the Dimensional Causality analysis.

Fig.\,\ref{fig:ASOMLearning} shows the development of the SOM during learning. At the beginning of the learning process, the $C_{i,j}$ centers of the receptive fields of the neurons are distributed uniformly in the unit cube. In each of the time steps, a random point of the manifold $Y(t)$ was chosen, and its local neighbourhood of $K=20$ points were presented to the network, forming a bundle in the manifold $X$ marked by different colors. As the learning proceeds, the predefined $20 \cdot 40$ grid structure of the network fits more and more smoothly to the embedded manifold of $X(t)$.

Fig.\,\ref{fig:ASOMReadout} A, shows a well-trained anisotropic 2D SOM that learned the topology of the manifold: the first dimension of the 2D map, shown by colored lines, fitted to extend along the submanifolds describing the self-dynamics, while the second dimension, marked by different colors spans across all submanifolds and corresponds to the values of the hidden common driver $Z$. Fig.\,\ref{fig:ASOMReadout} B shows the color-coded parts of the manifold $Z$ that correspond to the colored submanifolds on Fig.\,\ref{fig:ASOMReadout} A.

After training was completed, the $C_{i,j}$ weights of the network were frozen. A new data series of length $T=10000$ steps, generated using the same parameters but not used in the training process, was then used to evaluate the precision of the reconstruction. The discretized estimations of the hidden common driver were obtained from the network as the second index of the winner node. The reconstructed data was normalized by setting its mean to zero and its standard deviation to one (Fig.\,\ref{fig:ASOMReadout}, C, green line), and compared to the ground truth values of the hidden common driver, which were also normalized similarly (Fig.\,\ref{fig:ASOMReadout}, C, black line). Remarkably, the reconstructed hidden common driver closely follows the chaotic dynamics of the actual driver. To quantify the similarity, the correlation between the reconstructed $\hat z(t)$ and ground truth $z(t)$ values is presented (Fig.\,\ref{fig:ASOMReadout}, D). 

In this specific example, the linear correlation coefficient between the reconstructed and ground truth values reached $0.91$, while the $x(t)$ and $y(t)$ variables showed only instantaneous correlations of $0.24$ and $0.20$ with the hidden $z(t)$ variable, respectively. Interestingly, despite no time delay being introduced between the hidden common cause (HCC) and the driven systems (i.e., the effect was instantaneous), the maximum correlation coefficients were achieved with different delays: the maximum absolute value of the cross-correlation function between $x(t)$ and $z(t)$ was $0.42$, occurring with a 2-timestep delay, and $0.30$ between $y(t)$ and $z(t)$, occurring with a 6-timestep delay.

\begin{figure*}[htb!]
\centering
\includegraphics[width=\textwidth]{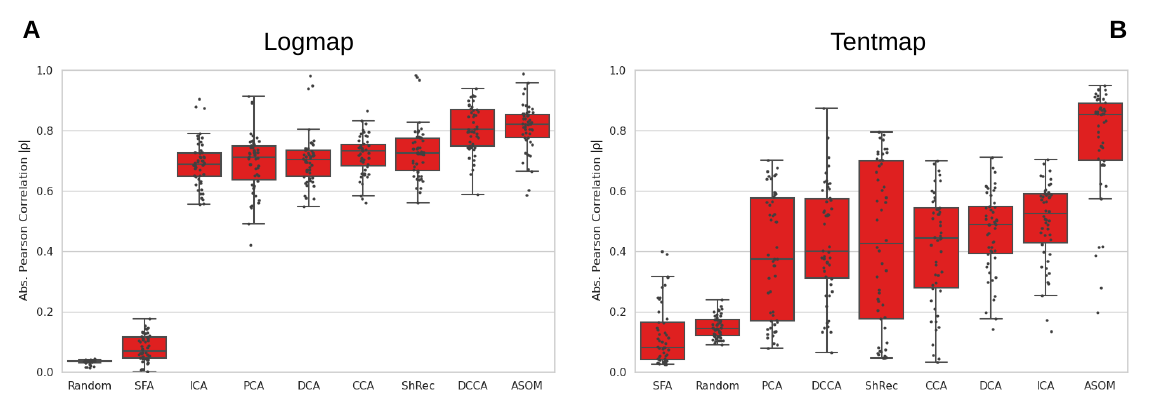}
\caption{\textit{Reconstruction performance comparisons.} The reconstruction performances are measured in terms of absolute correlation with the ground truth hidden common driver time series for 50 simulated coupled logistic map systems (A) and tent map systems (B). Individual results are shown by black dots, the median by solid black lines, the quadrilles are red squares, the 1.5 interquartille range by the whiskers and the outliers are with diamonds. The methods compared: SFA: Slow Feature Analysis, ICA: Independent Component Analysis, DCA: Dynamical Component Analysis, PCA: Principal Component Analysis, ShRec: Shared Recurrence \cite{Gilpin2023}, CCA: Canonical Correlation Analysis, DCCA: Deep Canonical Correlation Analysis, ASOM: Anisotropic Self-Orgainizing Map (our method). Our ASOM method showed superior precision in hidden common driver reconstruction in this test, compared to all the other methods examined for both dynamical systems.}
\label{fig:Comp}
\end{figure*}

The reliability of our method's performance was evaluated through repeated simulations with randomly chosen parameters. Two different dynamical systems were used in these tests: the coupled triad of logistic maps (Eq.\,\ref{eq:hcclogist}, similar to those used in previous examples) and the triad of coupled tent maps (Eq.\,\ref{eq:hcctent}).

The performance of the different methods was assessed based on the median and median absolute error of the absolute values of the correlation between the reconstructed and ground truth $z(t)$ values. These results are presented in Fig.\,\ref{fig:Comp} and the median absolute correlations are summarized in Tab.\,\ref{tab:result}. For both families of dynamical systems tested, our ASOM method achieved the highest median correlation values: $0.82$ for the logistic maps and $0.85$ for the tent maps.

Note that in our method, the hidden common driver values are estimated in a discretized manner using only 20 nodes. Therefore, the finite precision of the discretization imposes an upper limit on the correlation coefficient of 0.975.

\begin{table*}[htb!]
\centering
\setlength{\tabcolsep}{10pt}
\renewcommand{\arraystretch}{1.5}
\begin{tabular}{lccccccccc}
dataset & Random & SFA & ICA & DCA & PCA & ShRec & CCA & DCCA & ASOM \\
\hline
Logistic map & 0.04 & 0.07 & 0.69 & 0.70 & 0.71 & 0.73 & 0.73 & 0.80 & \bfseries 0.82 \\
Tent map & 0.14 & 0.08 & 0.52 & 0.49 & 0.37 & 0.42 & 0.44 & 0.40 & \bfseries 0.85\\
\end{tabular}
\caption{\textit{Performance comparison on two datasets.} Median absolute correlations $|\rho|$ between the actual and the estimated hidden common driver time-series for different methods.}
\label{tab:result}
\end{table*}

\section{Discussion}

In this paper, we introduced a novel approach to reconstruct hidden common driver dynamics based solely on observations of the driven systems. Our method builds on the theoretical foundations of Takens' and Stark's theorems. We estimate the intrinsic dimensions of the observed driven dynamical systems and determine the dimension of the common driver by calculating the mutual dimension of the observed systems. Using these dimension estimations, we set up a neural network model—a self-organizing map—to learn the attractor manifold structure of the embedded observed data through a new, anisotropic variant of Kohonen's learning algorithm.

The effectiveness of our ASOM method was demonstrated using coupled triads of two types of discrete-time chaotic dynamical systems: coupled logistic maps and coupled tent maps. In each case, two chaotic oscillators were driven by a third, similar one. We showed that, in a well-trained network, the second index of the actual winner node correlates well with the hidden common driver. After conducting multiple simulations, we concluded that our ASOM method provides more precise inference of the hidden common driver compared to several alternative methods. Although ASOM delivered only slightly better performance than Deep Canonical Correlational Analysis with the logistic maps, it significantly outperformed all other methods with the coupled tent map systems.

The dimension plays a central role in topological comparisons: bidirectional topographical mapping can only exist between manifolds of equal dimensions. CCM can result in asymmetric (unidirectional) relation only if the dimensions of the two manifolds differ. In this case, CCM implements a non-injective but surjective function from the higher dimensional consequence to the cause, but an injective but not surjective function from the cause to the consequence.

CCM implies unidirectional causal connection if the mapping is injective from the consequence to the cause but not in the reverse direction, from the cause to the consequence. As observed in case of the unidirectionally coupled logistic maps, when applying CCM from the cause to the consequence, a small neighborhood of a point in the cause manifold maps to a lower-dimensional submanifold in the consequence manifold, without covering the entire manifold. This causes problems in the evaluation of CCM results since the center of the submanifolds typically non-independent of the value of the cause, thus CCM correlation will be different from zero in this (reverse) direction (albeit smaller than one). As CCM in this reverse direction will converge to a value significantly different from zero, it does not allow for the application of statistical tests to draw clear conclusions.

Similar to the work of \cite{Sauer2004, Sauer2010} and \cite{Gilpin2023}, our approach exploits the property that the hidden common driver divides the manifold of the driven systems into submanifolds ie. equivalence classes, corresponding to the states of the hidden common driver. Sauer's method simply collects the points belonging to equivalence classes and chooses a representative from them. While it is straightforward to divide a 1D manifold into equivalence classes due to the proximity of points belonging to the same class, dividing a 2D manifold into 1D submanifolds as equivalence classes is more challenging. Our method solves this problem by fitting a neural grid of pre-defined dimensions. Our examples were among the simplest possible systems with 1D driver and 2D driven systems. Thus an ASOM with a 2D grid was enough to learn the dynamics of the driven system and the 1D driver dynamics within it. Systems with higher dimensions will pose further challenges, that require further studies.

If additional assumptions can be made about the nature of the hidden common driver, there are alternative methods for its reconstruction. For instance, Slow Feature Analysis (SFA) can reconstruct the hidden common driver based on the assumption that its dynamics evolve much more slowly than those of the driven systems \cite{Wiskott98, Wiskott2002}. In contrast, our approach does not rely on such assumptions. In our demonstration example, the chaotic dynamics of the hidden driver and the driven systems were all fast and shared similar characteristic frequencies.

Similarly, various statistical methods can be employed to approximate a hidden common driver under different assumptions. For example, Principal Component Analysis (PCA), Independent Component Analysis (ICA), and Dynamical Component Analysis (DCA) assume that the observed time series are linear combinations of the hidden variables. In contrast, our approach does not rely on such assumptions. In our demonstration example, the linear correlation between the observed and hidden systems was minimal and time-shifted.

One might reasonably assume that even non-linearly coupled hidden common driver time series could manifest in the hidden layers of deep autoencoders trained on the observed time series. However, identifying and extracting the relevant features from the numerous neurons in these hidden layers presents a significant challenge.

To address this issue, \cite{Benko2021} proposed a feed-forward deep neural network architecture called Mapper-Coatch. This approach uses error backpropagation to train the network to predict the future of one of the time series based on both observed dynamics. The estimation of the hidden common driver emerges from the information bottleneck between the two modules. Interestingly, an a priori estimation of the dimensionality of the hidden common driver is crucial in this approach as well. It is necessary to define the width of the bottleneck to achieve effective reconstruction.

\section*{Acknowledgements}
This research was supported by the Hungarian National Research, Development, and Innovation Office NKFIH, under grant numbers K113147 and K135837, the Hungarian Research Network HUN-REN under grant number SA-114/2021, and by the Hungarian National Brain Research Program  2017-1.2.1-NKP-2017-00002.

\section*{Declaration}
During the preparation of this work the authors used ChatGPT to improve readability and language. After using this tool, the authors reviewed and edited the content as needed and take full responsibility for the content of the publication.

\appendix

\section{Alternative models}
\label{sec:appendix}

\subsubsection*{Slow Feature Analysis (SFA)}

Slow Feature Analysis \cite{Wiskott98, Wiskott2002} was carried out on the preprocessed signals using the sksfa (version$=0.1.6$) python package.

We use the embedded signals from the observed variables purely as input features and set the number of components parameter to one. We applied the same procedure to the tent map dataset.

\subsubsection*{Principle Component Analysis (PCA)}

We applied PCA using the scikit-learn package \cite{scikit-learn}. For the logistic map and tent-map dataset we passed the embedded input signals from the observed subsystems and projected it to its fist principle component. We calculated the correlation between the projection and the hidden variable.

\subsubsection*{Independent Component Analysis (ICA)}

ICA transforms the multivariate signal into components by minimizing the mutual information between them to find good data representation. We used the ICA implementation from the scikit-learn package \cite{scikit-learn}.
A parameter of the method is the number of components on which the original high-dimensional signal is projected on. We set this parameter to $1$ for the logistic map and tent map datasets.

\subsubsection*{Dynamical Component Analysis (DCA)}
Dynamical Component Analysis is a linear dimensionality reduction technique that projects highdimensional time series into a subspace with maximal predictive information measured by mutual information \cite{Clark2019}. We used the official python implementation from github \cite{DCA}. DCA has two main parameters, the time window and the number of components used in the procedure. We set the time window parameter to $T=5$ and set the number of components to one for both datasets.

\subsubsection*{Canonical Correlation Analysis (CCA)}
We use Canonical Correlation Analysis to linearly project the observed subsytems to a subspace each where the correlation is maximized between the representations. We used the sklearn implementation of CCA \cite{scikit-learn}. For the logistic map and tent map datasets we embed the time series into $3$ dimensions and set the number of components to $1$. We get a projected value for each of the observed subsystems, compute the mean of the two, and use this value as an estimate for the hidden variable.

\subsubsection*{Deep Canonical Correlation Analysis (DCCA)}
We use Deep Canonical Correlation Analysis \cite{Andrew2013} to map the observed subsystems to 1D subspaces, where the correlation is maximized between the values. We used the DCCA implementation from the mvlearn python package \cite{perry2021mvlearn}. We apply the same data processing as in the case of linear CCA. We use multilayer feedforward neural networks with two hidden layers, with m=$20$ hidden units in each layer.

\subsubsection*{Shared Dynamics Reconstruction Algorithm (ShRec)}
We apply the shared dynamics reconstruction algorithm \cite{Gilpin2023} to recover the hidden driver behind the observed subsystems. We utilized the Python implementation on GitHub corresponding to the preprint. We set the embedding parameter to $3$ and leave all the other parameters on default values. We compare the reconstructed hidden variable to the predicted shared dynamics.

\subsubsection*{Phase shoufled baseline (Random)}
We create a random baseline for each dataset by applying a Fourier transformation to the hidden common driver time series. We then randomly shift the phase of each Fourier component and transform it back to the time domain. This process results in a phase-shuffled time series with the same spectrum as the original hidden common driver. This control is intended to reveal any correlation that might arise solely from similar spectra, which could be significant particularly in the low-frequency range.

 \bibliographystyle{elsarticle-num} 
 \bibliography{Biblio}





% \end{thebibliography}
\end{document}